%% file: main.tex
\documentclass{article}
\usepackage[margin=2.5cm]{geometry}
\usepackage{natbib}
\usepackage{hyperref}
\usepackage{enumitem}

\usepackage{amsthm}
\usepackage{hyperref}

\usepackage{amsfonts}

\newcommand{\CR}{\operatorname{CR}}
\newcommand{\SR}{\operatorname{SR}}
\newcommand{\Reg}{\operatorname{CR}}
\newcommand{\pai}{^{(i)}}
\newcommand{\Gap}{\text{Gap}}

\usepackage[dvipsnames]{xcolor}
\newcount\comments  
\newcommand{\genComment}[2]{\ifnum\comments=1{\textcolor{#1}{\textsf{\footnotesize #2}}}\fi}

\input{macros}

\title{The Fallacy of Minimizing Cumulative Regret in the Sequential Task Setting}

\author{
  Ziping Xu\\
  \texttt{Harvard University}
  \and
  Kelly Zhang\\
  \texttt{Imperial College London}\\
  \and
  Susan A. Murphy\\
  \texttt{Harvard University}
}

\begin{document}
\maketitle

\begin{abstract}
Online Reinforcement Learning (RL) is typically framed as the process of minimizing cumulative regret (CR) through interactions with an unknown environment. However, real-world RL applications usually involve a sequence of tasks, and the data collected in the first task is used to warm-start the second task. The performance of the warm-start policy is measured by \textbf{simple regret} (SR). While minimizing both CR and SR is generally a conflicting objective, previous research has shown that in stationary environments, both can be optimized in terms of the duration of the task, $T$.

In practice, however, in real-world applications, human-in-the-loop decisions between tasks often results in non-stationarity. For instance, in clinical trials, scientists may adjust target health outcomes between implementations. Our results show that task non-stationarity leads to a more restrictive trade-off between CR and SR. To balance these competing goals, the algorithm must explore excessively, leading to a CR bound worse than the typical optimal rate of $T^{1/2}$. These findings are practically significant, indicating that increased exploration is necessary in non-stationary environments to accommodate task changes, impacting the design of RL algorithms in fields such as healthcare and beyond.

\end{abstract}

\section{Introduction}
\input{sections/sec0_intro}

\section{Problem Setup}
\label{sec:setup}
\input{sections/sec1_setup}

\section{Minimax Lower Bound}
\label{sec:minimax}
\input{sections/sec3_minimax_new}

\section{Optimal Level of Exploration}
\label{sec:opt_expl}
\input{sections/sec4_opt_exploration}

\section{Study on Changes in $P$}
\label{sec:change_P}
\input{sections/sec5_changes_P}

\section{Extending to Multiple Non-linear Tasks}
\label{sec:multi-task}
\input{sections/sec6_extension}

\section{Simulation Studies}
\label{sec:simulation}
\input{sections/sec7_simulation}

\section{Discussion}
\label{sec:discussion}
In this paper, we study the trade-off between cumulative regret and simple regret across two contextual bandit tasks. We demonstrate that with non-stationarity between tasks due to human-in-the-loop decisions, the above trade-off is more restrictive. These changes include changes in policy spaces, reward mappings and outcome distributions, which is of significant novelty. A main message is that one should employ additional exploration compared to what is sufficient for single task cumulative regret minimization in presence of such changes. 

\paragraph{Limitations.} Potential limitations of this paper include the scope of theoretical results. As a first step towards the sequential multi-task setting with non-stationarity between tasks, there is a gap in the minimax rate presented in Theorem \ref{thm:1}. More specifically, the lower bound should depend on the complexity of the environment $|\gX|$ and $|\gA|$. Further work should be done to bring this bound to minimax-optimal.

We study minimax rate throughout the paper, which focuses often on the worst case. In reality, some instances are significantly harder to learn than the others. An interesting direction is to propose a theoretical measure of the significance of the trade-off studied in this paper and derive an instance-dependent result. 

\bibliographystyle{apalike}
\bibliography{main}

\newpage

\input{sections/appendix}

\end{document}

%% file: macros.tex
\usepackage{amsmath,bbm,bm}
\usepackage{mathtools}
\usepackage{algorithm}
\usepackage{algpascal}
\usepackage{algorithmicx}
\usepackage{algpseudocode}
\algdef{SE}[SUBALG]{Indent}{EndIndent}{}{\algorithmicend\ }%
\algtext*{Indent}
\algtext*{EndIndent}

\makeatletter
\newcommand*{\rom}[1]{\expandafter\@slowromancap\romannumeral #1@}

\newcommand\numberthis{\addtocounter{equation}{1}\tag{\theequation}}

\newtheorem{thm}{Theorem}
\newtheorem{lem}{Lemma}
\newtheorem{prop}{Proposition}

\newtheorem{rmk}{Remark}

\newtheorem{defn}{Definition}

\def\eqref#1{equation~\ref{#1}}

\def\1{\bm{1}}

\def\vy{{\bm{y}}}

\def\mP{{\bm{P}}}

\def\mS{{\bm{S}}}

\DeclareMathAlphabet{\mathsfit}{\encodingdefault}{\sfdefault}{m}{sl}
\SetMathAlphabet{\mathsfit}{bold}{\encodingdefault}{\sfdefault}{bx}{n}

\def\gA{{\mathcal{A}}}

\def\gF{{\mathcal{F}}}

\def\gH{{\mathcal{H}}}

\def\gL{{\mathcal{L}}}

\def\gO{{\mathcal{O}}}
\def\gP{{\mathcal{P}}}

\def\gS{{\mathcal{S}}}

\def\gX{{\mathcal{X}}}
\def\gY{{\mathcal{Y}}}

\def\sP{{\mathbb{P}}}

\def\sR{{\mathbb{R}}}

\def\sZ{{\mathbb{Z}}}

\newcommand{\E}{\mathbb{E}}

\DeclareMathOperator*{\argmax}{arg\,max}

%% file: sections/sec0_intro.tex
Cumulative regret (CR) minimization has emerged as a central topic in online Reinforcement Learning (RL) research \citep{auer2008near,chu2011contextual,agrawal2013thompson,lattimore2020bandit}. Within a stationary environment, prioritizing regret minimization is theoretically sound. A sublinear regret bound ensures a strong lower bound on cumulative rewards, as well as convergence to the optimal policy when the suboptimality gap is non-zero. However, any algorithm with a sublinear regret will stop exploration over time, which may lead to suboptimal performances for downstream tasks in a sequential task setting. 

\begin{figure}[hpt]
    \centering
    \includegraphics[width=0.8\textwidth]{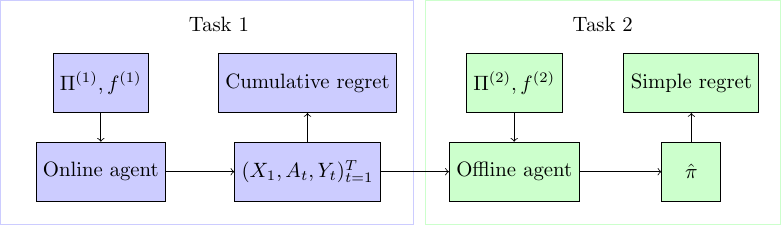}
    \caption{Two-task learning paradigm}
    \label{fig:paradigm}
\end{figure}

To explore the nature of this suboptimal performance, we consider a simple setting with two tasks, where the agent is restricted to deploy policies from a policy class $\Pi$ and the goal is to maximize rewards, defined as a mapping $f$ of outcome vectors. Stationarity holds across the two tasks if the reward mapping and policy class are the same.  However, often, due to human-in-loop decisions, the choice of the reward mapping and/or policy class changes between tasks. See Figure \ref{fig:paradigm} in which rewards are known functions of outcome vectors. In the first task, the reward is defined by the function $f^{(1)}$, and in the second task, the reward is defined by a different function $f^{(2)}$, where $f^{(2)}\neq f^{(1)}$.


\begin{enumerate}
    \item Task one: The agent observes the policy class $\Pi^{(1)}$, and the reward mapping $f^{(1)}$. It interacts with an unknown environment for $T$ steps. At each step $t$, the agent observes a context $X_t$ and samples an action $A_t \sim \pi_t(X_t)$ from $\Pi^{(1)}$. The environment then generates an outcome vector $Y_t$, and the agent aims to maximize the cumulative reward $\sum_t f^{(1)}(Y_t)$.
    \item Task two: The agent observes the policy class $\Pi^{(2)}$, and the reward mapping $f^{(2)}$. Based on the data collected during the online learning stage, it proposes a policy $\hat \pi \in \Pi^{(2)}$. The goal in this task is to maximize $\E[f^{(2)}(Y_t) \mid A_t \sim \hat \pi(X_t)]$.
\end{enumerate}

The focus on the fixed policy $\hat \pi$ for the second task is motivated by two key concerns. First, in real-world RL deployment, both safety and access to appropriate infrastructure are critical.  Indeed implementing an online RL algorithm (i.e., an adaptive policy) in practice presents challenges, such as building the necessary infrastructure to update models in near real-time \citep{pacchiano2024experiment}.
Additionally, deploying online RL algorithms often requires real-time monitoring systems \citep{trella2024monitoring} to mitigate negative impacts on users or human subjects \citep{liao2020personalized}, which can be costly. For these and other reasons, a fixed policy is often deployed in the subsequent tasks. 


We evaluate the performance of the task one and the task two through cumulative regret (CR) and simple regret (SR), respectively. Simple regret is the expected gap between the reward of the optimal policy in $\Pi^{(2)}$ and that of $\hat \pi$. \cite{krishnamurthy2023proportional} show the trade-off between $\CR$ and $\SR$ when the environment is stationary, i.e., $\Pi^{(1)} = \Pi^{(2)}, f^{(1)} = f^{(2)}$:
\begin{equation}
\label{equ:existing_lower}
    \inf_{\text{Algorithm}} \sup_{\text{Instance}} \E[\CR] \times\E[\SR] = \tilde{\Omega}(|\gX|\times|\gA|)
\end{equation}
for contextual bandits with context space $\gX$ and arm set $\gA$ \footnote{We presented a stronger version of Theorem 3 in \cite{krishnamurthy2023proportional} for a easier presentation.}. The above characterizes a weak trade-off between CR and SR in the sense that both CR and SR can be minimax-optimal with rates $\sqrt{|\gX||\gA| T}$ and $\sqrt{|\gX||\gA|/ T}$, respectively.


\paragraph{Human-in-the-loop between tasks.} The lower bound in (\ref{equ:existing_lower}) critically depends on the stationarity between tasks--the agent runs on the same environment across tasks. 
However, {\it many real-world RL tasks occur sequentially, with human-in-the-loop interventions between tasks that redefine the task parameters}.  This means that although the distribution of the outcome, $Y$, conditional on  context, action  $(X,A)$,    is the same across tasks, the policy class  changes, $\Pi^{(1)} \neq \Pi^{(2)}$ and/or the reward changes, $f^{(1)} \neq f^{(2)}$, leading to non-stationarity in the task.  
See, for example, applications in mobile health \citep{liao2020personalized,bidargaddi2020designing,trella2022designing}, and online education \citep{Aleven2023TowardsTF,Ruan2023ReinforcementLT}.

For example, in mobile health, an RL agent might initially deliver digital interventions based on the context $X_t$, such as sleep quality from the previous night. However, in a subsequent implementation, changes in user agreements may prevent the collection of $X_t$, requiring the agent to make decisions independently of that context. This results in a constrained policy class $\Pi^{(2)} = \{\pi: \pi(x_1) = \pi(x_2), \forall x_1, x_2\}$.



\paragraph{Main contribution.} We demonstrate that above changes between tasks, driven by human-in-the-loop decisions, can significantly worsen the trade-off between CR and SR compared to  the trade-off implied by (\ref{equ:existing_lower}). Specifically, we instantiate three types of task changes due to human-in-the-loop decisions and show the minimax rate of
$\sqrt{\E[\CR]} \E[\SR]  = {\Omega}(1)$ when these changes are present. This lower bound suggests a more restrictive trade-off between CR and SR than (\ref{equ:existing_lower}), since the optimal minimax cumulative regret rate $\sqrt{T}$, the typical rate achieved by most contextual bandit algorithms, results in a suboptimal rate for SR, $\E[\SR] = {\Omega}(T^{-1/4})$. This rate for SR is suboptimal because pure exploration in the first task gives $\E[\SR] = {\gO}(T^{-1/2})$. We further extend our results to settings with multiple tasks and validate the theory through simulation studies. 

The main message of this paper is that {\it additional exploration is required when unanticipated changes occur between tasks due to human-in-the-loop involvement. Complete exploitation, i.e. use of an algorithm leading to minimax-optimal cumulative regret, within a task can lead to worse performance in later tasks.}

%% file: sections/sec1_setup.tex
\paragraph{Notations.} For a set $\gX$, we denote by $\Delta(\gX)$ the set of all distributions over $\gX$. For $N \in \sZ$, we let $[N] = \{1, 2, \dots, N\}$. We use $\gO(\cdot)$, $\Theta(\cdot)$, $\Omega(\cdot)$ to denote the big-$O$, big-Theta and big-Omega notations. The $\tilde{\gO}(\cdot)$, $\tilde{\Theta}(\cdot)$, $\tilde{\Omega}(\cdot)$ notation hides all the logarithmic terms. We denote by $D_{\operatorname{KL}}(P \mid Q)$, the KL divergence between two probability measures $P$ and $Q$ with $P \ll Q$.

\paragraph{Two-task contextual bandit learning paradigm.} 
We rigorously introduce the two-task contextual bandit paradigm in Figure \ref{fig:paradigm}. The agent operates in a contextual bandit environment with context space $\gX$, action set $\gA$, and outcome space $\gY$. Here we consider discrete set $\gX$. We denote by $\Pi$ the set of all mappings from $\gX$ to $\Delta(\gA)$, i.e., Markovian policies.

In the first task, the agent is given a policy space $\Pi^{(1)} \subset \Pi$ and reward mapping $f^{(1)}: \gY \mapsto \sR$. At each step $t \in [T]$, the environment generates a context $X_{t} \in \gX$ i.i.d. from the context distribution $P_X$. The agent chooses a policy $\pi_t: \gX \mapsto \Delta(\gA)$, and samples an action $A_t \sim \pi_t(X_t)$, where $\pi_t \in \Pi^{(1)}$. The environment generates a random outcome vector $Y_{t} \sim P(X_t, A_t)$, with outcome distribution $P: \gX \times \gA \mapsto \Delta(\gY)$. The reward during task 1 is $R_t = f^{(1)}(Y_t)$. The agent minimizes cumulative regret in Definition \ref{defn:cum_regret}.

\begin{defn}[Cumulative regret]
\label{defn:cum_regret}
    Denote by $R_{\pi}^{(1)} \coloneqq \E_{X_t \sim P_X, A_t \sim \pi(X_t)}\E_{Y_t \sim P(\cdot \mid x, a)} f^{(1)}(Y_t)$ the mean reward of a policy $\pi$. The cumulative regret is  
$$
    \CR \coloneqq \sum_{t = 1}^{T} \left[\max_{\pi \in \Pi^{(1)}} \E[R^{(1)}_{\pi} - R^{(1)}_{\pi_t}] \right].
$$
\end{defn}

For the second task, the agent is given a new policy space $\Pi^{(2)} \subset \Pi$ and reward mapping $f^{(2)}: \gY \mapsto \sR$. It then learns offline from the data collected during task 1 and proposes a policy $\hat \pi \in \Pi^{(2)}$ . The goal is to minimize simple regret $\SR(\hat \pi)$.

\begin{defn}[Simple regret]
\label{defn:simple_regret}
Define the simple regret of a policy $\pi$ by
$$
    \SR(\pi) \coloneqq \max_{\pi' \in \Pi^{(2)}} R_{\pi'}^{(2)} - R_{\pi}^{(2)},
$$
where $R_{\pi}^{(2)} \coloneqq \E_{X_t \sim P_X, A_t \sim \pi(X_t)}\E_{Y_t \sim P(\cdot \mid x, a)} f^{(2)}(Y_t)$.
\end{defn}

Note that both $\CR$ and $\SR$ depend implicitly on the learning algorithm and the environment parameters including $(P, \Pi^{(1)}, \Pi^{(2)}, f^{(1)}, f^{(2)})$. So far we consider \textbf{the same outcome distribution} $P$ to focus on the study of changes in policy spaces and reward functions. We extend our discussion to outcome distribution shift in Section \ref{sec:change_P}.

\begin{rmk} Policy space and reward mappings $\Pi^{(i)}, f^{(i)}$ are part of the task design and are therefore assumed to be known before the agent interacts with the $i$-th task. The agent is learning the underlying outcome distribution $P$, that is considered unknown, and remains the same for two tasks.
\end{rmk}

\paragraph{Motivations for changes between tasks.} The setup for the two tasks may change in various ways. Below are motivating examples of changes in $(\Pi\pai, f\pai)$.
\begin{enumerate}
    \item \textbf{Changes in reward mappings $f\pai$.} In mobile health studies, we observe two outcomes $Y_t = (Y_{t, 1}, Y_{t, 2})$, where $Y_{t, 1}$ is the indicator of healthy behaviors, and $Y_{t, 2}$ is the response rate. During the online learning, the reward mapping is $f^{(1)}(Y_{t}) \equiv Y_{t, 1}$. Based on the observed data, a domain expert may decide that the response rate is too low, which could deter participants from using the app. Therefore, they propose a new reward mapping $f^{(2)}(Y_t) = Y_{t, 1} - \alpha Y_{t, 2}$, where $\alpha$ is a balancing factor.
    \item \textbf{Changes in policy classes $\Pi\pai$.} In real-world applications, the context space $\gX$ can be a high-dimensional vector due to the complexity of real-world observations. During online learning, computational limitations may prevent optimization over the entire space of policies that account for all contexts. Hence, the domain expert might decide to optimize only in the space of context-independent policies $\Pi^{(1)} = \{\pi \in \Pi: \pi(a \mid x_1) = \pi(a \mid x_2), \text{ for all } (x_1, x_2) \in \gX\}$. As more data is collected, the expert may decide that certain components of $\gX$ are relevant to the task, which should be included in the offline learning. 
\end{enumerate}

%% file: sections/sec3_minimax_new.tex

A major result of this paper is the more restrictive trade-off between cumulative regret and simple regret. We demonstrate this trade-off by establishing a minimax lower bound under mild conditions on the problem instance set.

\paragraph{Problem instance.} Tuple $(P, \Pi^{(1)}, \Pi^{(2)}, f^{(1)}, f^{(2)})$ uniquely defines the environment the algorithm in which the algorithm operates. Since policy classes and reward mappings are known, we denote a problem instance simply by $P$, the unknown outcome distribution, and by $\gP$ the set of outcome distributions of interest.

\begin{defn}[Occupancy measure]
    Given a policy $\pi$, we define the occupancy measure of $\pi$ as 
    $$
        \mu_{\pi}(x, a) = P_X(x) \pi(a \mid x).
    $$
    Note that occupancy measure is independent of bandit instance as we assume the same context distribution.
\end{defn}

\begin{defn}[Learning algorithm]
    We denote a learning algorithm by $L = (L^{(1)}, L^{(2)})$, where $L^{(1)}$ is an online learning algorithm that is a deterministic mapping from history $\gH_t = (X_{\tau}, A_{\tau}, Y_{\tau})_{\tau = 1}^{t-1}$ to a policy in $\Pi^{(1)}$. Here, $L^{(2)}$ is an offline learning algorithm that maps $\gH_{T+1}$ to a policy $\hat \pi \in \Pi^{(2)}$. Note that $L^{(i)}$ can depend on $f^{(i)}$ and $\Pi^{(i)}$ for each $i \in \{1, 2\}$ as they are known knowledge. We denote the set of all such algorithms by $\gL$.
\end{defn}

Now we are ready to present our main theorem. 
\begin{thm}
    \label{thm:1}
    For the policies spaces $\Pi^{(1)} = \{\pi: \pi(\cdot \mid x_1) = \pi(\cdot \mid x_2), \text{for all } x_1, x_2 \in \gX\}$ and $\Pi^{(2)} = \Pi$, there exists an instance set $\gP$, and reward mapping $f^{(1)} = f^{(2)}$, such that, 
    \begin{equation}
        \inf_{L \in \gL} \sup_{P \in \gP} \E[\SR(\hat \pi)] \sqrt{\E[\CR]} = \Omega(1). \label{equ:thm11}
    \end{equation}
    Similarly, for some $f^{(1)} \neq f^{(2)}$, there exists exists an instance set $\gP$ and $\Pi^{(1)} = \Pi^{(2)} = \Pi$, such that 
    \begin{equation}
        \inf_{L \in \gL} \sup_{P \in \gP} \E[\SR(\hat \pi)] \sqrt{\E[\CR]} = \Omega(1). \label{equ:thm12}
    \end{equation}
\end{thm}

\paragraph{Discussion.} Theorem \ref{thm:1} demonstrates a more restrictive trade-off between cumulative regret and simple regret than the trade-off shown in (\ref{equ:existing_lower}) \citep{Athey2022ContextualBI}. This more restrictive result arises from allowing either $\Pi^{(1)} \neq \Pi^{(2)}$ or $f^{(1)} \neq f^{(2)}$.

A significant difference between (\ref{equ:thm11}), (\ref{equ:thm12}) and the existing result in (\ref{equ:existing_lower}) is that the rate for $\E[\CR]$ in (\ref{equ:existing_lower}) becomes square root. To understand the rationale of the square root term, we note that a purely random exploration in the first task results in $\E[\CR] = \Theta(1)$, which leads to a simple regret bound of $\E[\SR] = \Omega(1/\sqrt{T})$. This is consistent with the regular estimation error lower bound for estimating parametric models.

\subsection{Lower bound construction}

In this section, we discuss the lower bound construction and provide the intuition for the lower bound in Theorem \ref{thm:1}.

The overall idea is to construct instances where s specific context-arm pair $(x, a)$ leads to non-zero regret in task 1 whenever $(x, a)$ is visited. In this setting, $\mu_{\pi^*_1}(x, a) = 0$, meaning that the optimal policy for task 1 avoids visiting $(x, a)$ for cumulative regret minimization. However, in task 2, the same pair $(x, a)$ becomes optimal ($\mu_{\pi^*_2}(x, a) > 0$), and the offline learning algorithm will need a dataset with sufficient coverage on $(x, a)$ to learn the optimal policy.

This motivation is closely related to the results in the offline learning literature, where it has been shown that offline learning is fundamentally hard if the single-policy concentrability defined by $\max_{x, a} \mu_{\pi_1^*}(x, a) / \mu_{\pi_2^*}(x, a)$ is unbounded \citep{chen2019information}.

Based on this motivation, we now describe the detailed lower bound construction.

\paragraph{Case of $\Pi^{(1)} \neq \Pi^{(2)}$: add a new feature.} 

During the online learning, we may not have the computational resources to maximize over the space of all policies that takes context into account. As a result, the domain expert might decide to optimize only in the space of context-independent policies, turning the problem into a multi-armed bandit problem. With evidence accumulating, the expert may decide that certain components of $\gX$ are relevant to the task, which should be included in the offline learning. 

To illustrate this, we consider a set of two-armed contextual bandits with context space $\gX = \{x_1, x_2\}$ and outcome space $\gY = \sR$. We let $\Pi^{(1)} = \{\pi: \pi(\cdot \mid x_1) = \pi(\cdot \mid x_2), \text{for all } x_1, x_2 \in \gX\}$, meaning the policies do not differentiate between contexts. For task 2, the policy class is the set of all policies. We consider $f^{(1)}(y) = f^{(2)}(y) = y$. That is, the reward mappings are identical mappings.


We consider an instance set (outcome distribution set) $\gP$ such that the mean of $P(\cdot \mid x, a)$ for each context-arm pair is given by Table \ref{tab:case1_main} for any $\epsilon \in [0, 0.5]$ and $\xi \in [0, 0.25]$. Each choice of $\epsilon, \xi$ defines two instances, denoted by $P_{\epsilon, \xi}$, $\bar P_{\epsilon, \xi}$. We introduce $\xi$ to ensure the richness of the instance class. 

Note that in task 1 the algorithm must ignore context, thus it pulls the arm with the larger marginal mean. In all $P \in \gP$, the marginal mean reward for $a_1$ is strictly larger than that of $a_2$. This ensures that online learning algorithm must avoid pulling $a_2$, meaning that they will not collect enough data to distinguish between $P_{\epsilon, \xi}$ from its counterpart $\bar P_{\epsilon, \xi}$. 

However, in task 2, distinguishing between $P_{\epsilon, \xi}$ and $\bar P_{\epsilon, \xi}$ becomes crucial for minimizing simple regret because these two distributions have opposite optimal policies. As a result, we create a fundamental tension between cumulative regret and simple regret: minimizing cumulative regret in task 1 leads to poor data collection, making it difficult to minimize simple regret in task 2.

\vspace{-3mm}

\begin{table}[hpt]
    \centering
    \caption{Mean reward for $\Pi^{(1)} \neq \Pi^{(2)}$ case}
    \label{tab:case1_main}
    \begin{tabular}[t]{c|c|c|c}
        \hline
        \hline
        $P_{\epsilon, \xi}$ & $x_1$ & $x_2$ & marginal  \\
        \hline
        $a_1$  & $0.5+\epsilon$ & $0.5-\epsilon$ & 0.5 \\
        $a_2$ & $0.5-2\xi$ & 0.5 & $0.5-\xi$ \\
        \hline
        $\bar{P}_{\epsilon, \xi}$ & $x_1$ & $x_2$ & marginal \\
        \hline
        $a_1$ & $0.5+\epsilon$ & $0.5-\epsilon$ & 0.5  \\
        $a_2$ & $0.5-2\xi$ & $0.5-2\epsilon$ & $0.5-\xi-\epsilon$
\end{tabular}
\end{table}

\vspace{-3mm}

\paragraph{Case of $f^{(1)} \neq f^{(2)}$: change reward mapping.} 

We consider multi-armed bandit problems with no context. The outcomes vectors are $Y_t = (Y_{t, 1}, Y_{t, 2})$. The reward mappings are given by $f^{(1)}(Y_t) = Y_{t, 1}$ for task 1 and $f^{(2)}(Y_t) = Y_{t, 2}$ for task 2.

Consider the instance set $\gP$ such that the means of $Y_{t, 1}$ and $Y_{t, 2}$ for each arm pair is given in Table \ref{tab:case2_main} for any $\epsilon \in [0, 0.5]$, and $\xi \in [0, 0.25]$. Each choice of $\epsilon$ and $\xi$ defines two instances, denoted by $P_{\epsilon, \xi}$, $\bar P_{\epsilon, \xi}$. 

In this setup, it is clear that $a_2$ is the optimal arm in task 1, while the algorithm must pull $a_2$ to distinguish $P_{\epsilon, \xi}$ from $\bar P_{\epsilon, \xi}$, which is critical to decide the optimal policy for the second task. 

\begin{table}[hpt]
    \caption{Mean reward for $f^{(1)} \neq f^{(2)}$ case}
    \label{tab:case2_main}
    \centering
    \begin{tabular}[t]{c|c|c || c | c|c}
        $P_{\epsilon, \xi}$ & $Y_{t, 1}$ & $Y_{t, 2}$ & $\bar P_{\epsilon, \xi}$ & $Y_{t, 1}$ & $Y_{t, 2}$ \\
        \hline
        $a_1$  & 0.5 & 0.5 & $a_1$  & 0.5 & 0.5\\
        $a_2$  & 0.5-$\xi$ & 0.5-$\epsilon$  & $a_2$ & 0.5-$\xi$ & 0.5+$\epsilon$
    \end{tabular}
\end{table}

%% file: sections/sec4_opt_exploration.tex
As implied by Theorem \ref{thm:1}, any algorithm that achieves an optimal rate in $\CR$ will be suboptimal in $\SR$ in terms of the dependence on $T$. To balance between these two objectives, incorporate additional exploration during the first task. In this section, we characterize the optimal level of additional exploration that minimizes the following weighted objective for different values of $T'$:
\begin{equation}
    \E[\CR] + T' \E[\SR], \label{equ:weighted_sum}
\end{equation}
where $T'$ could be interpreted as the number of steps intended to be taken in task 2. Since we primarily focus on the role of horizons, we omit the dependence on $|\gX|$ and $|\gA|$ throughout the discussions in this section.

Proposition \ref{prop:lower_bound_sum} suggests a minimax lower bound for the weighted sum of cumulative regret and simple regret in (\ref{equ:weighted_sum}) that is the maximum of three terms: $T'/\sqrt{T}, T'^{2/3}$ and $\sqrt{T}$. The first term $T' / \sqrt{T}$ corresponds to the case where simple regret $T'\E[\SR]$ dominates, and the minimax rate of simple regret in the second for any dataset collected during the first task is $T' / \sqrt{T}$. The second term $T'^{2/3}$ corresponds to the rate characterized by Theorem \ref{thm:1}. The last term of $\sqrt{T}$ is the minimax rate of the first task regret minimization. This corresponds to the case when cumulative regret $\E[\CR]$ dominates.
\begin{prop}
\label{prop:lower_bound_sum}
Following the same choice of the instance set $\gP$ and $\Pi^{(1)}, \Pi^{(2)}, f^{(1)}, f^{(2)}$ in Theorem \ref{thm:1}, the following minimax lower bound holds
\begin{equation}
    \inf_{L \in \gL} \sup_{P \in \gP} \E[\CR + T'\SR] = \Omega\left( \max\left\{\frac{T'}{\sqrt{T}}, T'^{2/3}, \sqrt{T}\right\}\right).\label{equ:three_lower}
\end{equation}
\end{prop}

We show that a simple algorithm that mixes a minimax-optimal online learning algorithm with a purely random exploration policy has upper bounded $\CR$ and $\SR$ that matches the lower bound in Theorem \ref{thm:1} up to a factor of $|\gX| |\gA|$. This also allows us to achieve any point on the Pareto frontier up to a factor of $|\gX| |\gA|$. The parameter $\alpha$ controls the level of additional exploration in the first task. 
\begin{thm}
\label{thm:upper_bound}
    Let $L_0$ be an online learning algorithm with a regret bound of $\tilde{\gO}(\sqrt{|\gX||\gA|T})$. Let the algorithm for the first task be $L_{\alpha}(\tau) = (1-\alpha) L_0(\tau) + \alpha \pi_0$, where $\tau$ is any past observations and $\pi_0$ is the uniform random policy. For any choice of $\alpha \in [{|\gX| |\gA|}/{\sqrt{T}}, 1]$, there exist offline-learning algorithm for the second task such that
    \begin{equation}
        \E[\CR] = \tilde{\gO}\left(\alpha T \right), \E[\SR] = \tilde{\gO}\left(\sqrt{\frac{(|\gX| |\gA|)^2}{\alpha T}} \right). \label{equ:upper_bound}
    \end{equation}
\end{thm}

By tuning the exploration rate $\alpha$, we are able to match the minimax lower bound in (\ref{equ:three_lower}). In short, there are three regimes of $(T, T')$, for which we should choose different levels of exploration rate $T'$ to balance $\E[\CR]$ and $\E[\SR]$.

\begin{enumerate}
    \item \textbf{Regime 1: $T \leq T'^{2/3}$.} The first task is too short compared to the weight $T'$ of the second task, and the algorithm should employ pure exploration in the first task ($\alpha = 1$). This regime leads to a global regret of $\gO({T'}/{\sqrt{T}})$.
    \item \textbf{Regime 2: $T'^{2/3} < T \leq T'^{4/3}$.} The algorithm should employ additional exploration compared to these that achieve a minimax optimal rate in a single task. Theorem \ref{thm:upper_bound} suggests an additional exploration rate of $\alpha = T'^{2/3} / T$ and a global regret bound of $\gO(T'^{2/3})$. Note that under a special case of $T' = T$, the rate of $\alpha = T^{-1/3}$ indicates a regret bound of $\gO(T^{2/3})$ in the first task.
    \item \textbf{Regime 3: $T > T'^{4/3}$.} The algorithm should employ $\alpha = 0$, meaning that no excess exploration is needed and the agent in task 1 can minimize the cumulative regret as much as possible. In this regime, the cumulative regret in the first task could achieve the minimax optimal rate of $\sqrt{T}$. 
\end{enumerate}

It is often in real-world applications that $T'$ is pre-determined and the researcher could decide how many samples to collect in the first task to ensure a good learning in the second one. For instance, in an inventory management context \citep{madeka2022deep}, it is determined by the engineering team that how long a learned policy should be deployed for the second task. In such cases, our theory indicates that one should choose $T > T'^{4/3}$, so a greedy cumulative regret minimization for the first task is justified.

%% file: sections/sec5_changes_P.tex
In real-world implementations, the outcome distribution $P$ often experiences unpredictable shift. Prior research on non-stationary bandits has typically focused on single-task scenarios with potential reward distribution shifts at any step. To manage these shifts, the literature often limits the total variation in distribution shifts, making it possible to establish sublinear regret bounds. However, in a sequential task setting, when the algorithm cannot adaptively learn during the second task, the simple regret is always lower bounded by a constant. This is due to the uncertainty of the optimal policy for the second task, even with full knowledge of the outcome distribution from the first task. To address this challenge, we introduce the concept of {\it robust simple regret}. We show that the robust simple regret are shown to have a minimax lower bound similar to that shown in Theorem \ref{thm:1}.

For simplicity, we consider no change in the policy space or the reward function. Specifically, let $\Pi^{(1)} = \Pi^{(2)} = \Pi$, the set of all policies, and $f^{(1)} = f^{(2)} = f$, the identical mapping in $\sR$. We denote by $P^{(1)}$ and $P^{(2)}$ the outcome distributions, aka., reward distribution of the first and second task. We denote a problem instance by $\mP = (P^{(1)}, P^{(2)})$.

In this setting, the adversary is allowed to choose $P^{(2)}$ from a $L_1$ ball around $P^{(1)}$. This leads to the instance set $\gP(\Delta)$ parametrized by constant $\Delta$ such that each $\mP = (P^{(1)}, P^{(2)})$ satisfies 
\begin{align*}
P^{(2)} \in \gP(P^{(1)}, \Delta) \coloneqq 
     \left\{P: \sum_{a} |P(x, a)- P^{(1)}(x, a)| \leq \Delta \text{ for all } x \in \gX\right\}, \numberthis
\end{align*}
where we abuse the notation for $P$ and let $P(x, a)$ denote the mean reward for $(x, a)$.

\paragraph{Distributionally robust simple regret.} When $P^{(2)}$ is allowed to change adversarially from $P^{(1)}$, it is not reasonable to compare against the true optimal policy with respect to the unknown $P^{(2)}$. Instead, we define a robust regret notion. First we define the worst-case simple regret of a policy $\pi$: 
\begin{align*}
\SR(\pi \mid P^{(1)}, \Delta) \coloneqq
    \sup_{\substack{P^{(2)} \in \gP(P^{(1)}, \Delta) \\  x \in \gX}} \left(\max_{a} P^{(2)}(x, a) - \sum_{a} P^{(2)}(x, a) \pi(a \mid x)\right).
    \numberthis
\end{align*}
We denote by $\tilde{\pi}_{P^{(1)}, \Delta} \coloneqq \inf_{\pi' \in \Pi^{(2)}}\SR(\pi' \mid P^{(1)}, \Delta)$ the optimal robust policy given $(P^{(1)}, \Delta)$. When it is clear from the context, we drop the subscription for $P^{(1)}$ and $\Delta$.

We further define {\it robust simple regret}, which is the gap between the worst-case simple regret of a given policy and the policy that achieves the lowest worst-case simple regret: 
\begin{align*}
    \widetilde{\SR}(\pi \mid P^{(1)}, \Delta) \coloneqq 
    \SR(\pi \mid P^{(1)}, \Delta) - \inf_{\pi' \in \Pi^{(2)}}\SR(\pi' \mid P^{(1)}, \Delta).
    \numberthis
\end{align*}

Note that the worst-case regret form over some ambiguity set has been studied in the robust Markov Decision Process literature \citep{xu2010distributionally,eysenbach2021maximum,dong2022online}. However, the definition of robust simple regret and the tension between cumulative regret and robust simple regret has not yet been explored.

To illustrate how the trade-off between cumulative regret and simple regret remains relevant in the robust setting, we investigate a simple two-armed, context-free bandit case in Proposition \ref{prop:simple-regret}. The optimal arm in the the first task is $a_1$, while the optimal robust policy depends on the gap between the mean reward of both arms. Thus, to reduce the robust simple regret in the second task, the algorithm is forced to have an accurate estimate on the suboptimal arm in the first task.

\begin{prop}
\label{prop:simple-regret}
    Consider the following two-armed, context-free bandit, with $\Gap \coloneqq P^{(1)}(a_1) - P^{(1)}(a_2) > 0$. Then the worst-case simple regret is given by
    \begin{equation}
        \SR(\pi \mid P^{(1)}, \Delta) = \max\{  (\Delta - \Gap) \pi(a_1), (\Delta + \Gap) \pi(a_2)\}.
    \end{equation}
    Assume that $\Delta > \Gap$. The optimal robust policy $\tilde{\pi}$ w.r.t. the worst-case simple regret has the explicit form
    \begin{equation}
        \tilde{\pi}_{P^{(1)}, \Delta}(a_1) = \frac{\Delta + \Gap}{2\Delta}, \text{ and } \tilde{\pi}_{P^{(1)}, \Delta}(a_2) = \frac{\Delta - \Gap}{2\Delta}.
    \end{equation}
\end{prop}

Motivated by the instance introduced in Proposition \ref{prop:simple-regret}, we show the following Theorem that lower bounds the minimax rate of the product between cumulative regret in the first task and the robust simple regret in the second task. Note that robust simple regret does not depend on the actual $P^{(2)}$ of choice, the supremum is only taken over $P^{(1)}$.

\begin{thm}
\label{thm:lower_bound_p}
Assume each $P^{(i)}(\cdot \mid x, a)$ is from a Bernoulli distribution with mean $P^{(i)}(x, a)$ for all $i = 1, 2$ and $x, a \in \gX \times \gA$. Let $\gP^{(1)}$ be the set of such distributions. Then there exists some $\Delta$ such that 
\begin{equation}
    \inf_{L \in \gL} \sup_{P^{(1)} \in \gP^{(1)}}  \E[\widetilde{\SR}(\hat \pi \mid P^{(1)}, \Delta)] \sqrt{\E[\CR]} = \Omega(1).
\end{equation}
\end{thm}

Theorem \ref{thm:lower_bound_p} implies that one should still employ additional exploration for small $T$, when there is only changes in the outcome distributions.

%% file: sections/sec6_extension.tex
In our two-task study, we highlighted the inherent dilemma between cumulative regret and simple regret. Now we extend our discussion to a sequence of multiple tasks. A significant property about the two-tasks scenario is algorithm's inability to adaptively learn in the second task. This restriction forces the algorithm to "over-explore" in the first task to propose a good policy for the second task, thereby introducing a tension between the regrets in the first and the second task. In fact, such tension can exists between any task $i \in [N]$ and its preceding tasks over a sequence of $N$ tasks in the case that the algorithm is not allow to adaptively learn in all tasks. In this multi-task setting with no adaptivity, we demonstrate that a typical UCB type algorithm is fallacious leading to constant regret over an exponentially long sequence of tasks when the underlying environment is a nonlinear bandit.

\paragraph{Nonlinear contextual bandit.} For simplicity, we consider the outcome $\gY = \sR$ and a fixed reward function $f^{(i)} = f \equiv x \mapsto x$ for all tasks $i$, i.e. no rich observations, so we focus on the changes in the policy spaces. Following the setups in Section \ref{sec:setup}, we now consider potentially large or continuous context and action space $\gX$ and $\gA$. We define the mean reward as $R(x, a) \coloneqq \E_{Y \sim P(x, a)}[f(Y)]$. For contextual bandit with nonlinear reward models, we assume that mean reward $R \in \gF$ for some known function class $\gF: \gX \times \gA \mapsto \sR$. 

\paragraph{Complexity for nonlinear bandit.} Running UCB on nonlinear bandit is generally hard. \cite{russo2013eluder} proposed to explore by choosing
$
    A_t \in \argmax_{a \in \gA} \sup_{f \in \gF_t} f(X_t, a),
$
where $\sup_{f \in \gF_t} f(a)$ is an optimistic estimate of $f_{\theta}(a)$. A choice of $\gF_t$ given by \cite{russo2013eluder} is 
\begin{equation}
    \gF_t  = \left\{f \in \gF: \|f - \hat f_{t}^{LS}\|_{2, E_t} \leq \sqrt{\beta_t^\star}\right\}, \label{equ:F_t}
\end{equation}
where $\beta_t^\star$ are constants, $\|g\|_{2, E_t} = \sum_{t = 1}^T g^2(X_t, A_t)$ is the empirical 2-norm, and $\hat f_{t}^{LS} \in \inf_{f \in \gF} (f(X_t, A_t) - Y_t)^2$ is the empirical risk minimizer. The regret of running UCB with appropriately chosen $\beta_t^\star$ has regret of $\sqrt{\operatorname{dim}_{E}(\gF, T^{-2}) T}$, where $\operatorname{dim}_{E}(\gF, T^{-2})$ is the eluder dimension of the function class $\gF$.
\begin{defn}[Distributional eluder dimension]
    Let $\gF: \gX \mapsto \sR$. A probability measure $\nu$ over $\gX$ is said to be $\epsilon$-independent of a sequence of probability measures $\{\mu_1, \dots, \mu_n\}$ w.r.t $\gF$ if any pair of functions $f, \bar f \in \gF$ satisfying $\sqrt{\sum_{i = 1}^n(\E_{\mu}[f(x) - \bar f(x)])^2} \leq \epsilon$ also satisfies $|\E_{\nu}[f(x) - \bar f(x)]| \leq \epsilon$. Furthermore, $x$ is $\epsilon$-independent of $\{\mu_1, \dots, \mu_n\}$ if it is not $\epsilon$-dependent of the sequence.

    The $\epsilon$-eluder dimension $\operatorname{dim}_E(\gF, \epsilon)$ is the length of the longest sequence of distributions over $\gX$ such that for some $\epsilon' \geq \epsilon$, every distribution is $\epsilon'$-independent of its predecessors.
\end{defn}

Recall that the construction of our hard instances in Theorem \ref{thm:1} requires that the new task has the optimal policy whose occupancy measure has no overlap from the occupancy measure of optimal policies in the previous tasks. A generalization of this to the nonlinear case is that a predicted function that minimizes the loss over the dataset collected in the previous tasks may still occur large loss on a new task. Let the optimal policies of tasks $1, \dots n$ be $\pi_1^\star, \dots, \pi_n^\star$. Intuitively, as long as $n$ is smaller than $\operatorname{dim}_E(\gF, \epsilon)$, we can find a new task with optimal policy $\pi_{n+1}^\star$ for which the occupancy measure $\mu_{\pi_{n+1}^\star}$ is $\epsilon$-independent of $(\mu_{\pi_1^\star}, \dots, \mu_{\pi_n^\star})$. By the definition of eluder dimension, this implies that the function chosen for task $n+1$ based on the dataset collected by $(\mu_{\pi_1^\star}, \dots, \mu_{\pi_n^\star})$ may still occur a large error. Note that by running a no-regret online algorithm, the dataset collected during a task will asymptotically distributed as the occupancy measure induced by its optimal policy.

Eluder dimension has been shown to be exponentially large for simple models like one-layer neural network with ReLU activation function \citep{dong2021provable}. It is not trivial to show a lower bound directly depending on the eluder dimension. Instead, we provide a concrete example, where UCB described in (\ref{equ:F_t}) fails.

\begin{thm}
    Consider the hypothesis set $\gF$ to be one-hidden layer neural networks with width $d$. There exists ground-truth reward function and a sequence of tasks of length $\Omega(\exp(d))$ with different $\Pi^{(i)}$, such that the cumulative regret for each task is lower bounded by a constant, even if each $T_i \rightarrow \infty$.
    \label{thm:UCB}
\end{thm}

Theorem \ref{thm:UCB} indicates that even without a change in outcome distributions, there still exists an exponentially long sequence of tasks, for which the tension between local regret minimization and global regret minimization still holds. An UCB algorithm that greedily minimizes local regret fails to provide good guarantees for later tasks.

%% file: sections/sec7_simulation.tex
A main result of this paper is that the trade-off characterized in (\ref{equ:existing_lower}) can be significantly more restrictive when certain changes present. We validate this theory with two contextual bandit experiments, one focusing on the policy classes change and the other focusing on the reward mappings change. The parameters are chosen similarly as the hard-instance construction discussed in Section \ref{sec:minimax} to ensure the tension between cumulative regret and simple regret. Specifically, in experiment 1, we choose $\Pi^{(2)} = \Pi$ and $\Pi^{(1)} = \{\pi: \pi(\cdot \mid x_1) = \pi(\cdot \mid x_2), \text{for all } x_1, x_2 \in \gX\}$. In experiment 2, we choose $f^{(1)}(\vy) = y_1$ and $f^{(2)}(\vy) = y_2$ for $\vy = (y_1, y_2)$. 

For each experiment, we run UCB \citep{auer2002using} and UCB mixed with probability 0.1 random exploration. We compute the average cumulative regret and simple regret over time.

\begin{figure}[hpt]
\centering
\begin{minipage}[b]{1\textwidth}
\centering
    \includegraphics[width = 0.35\textwidth]{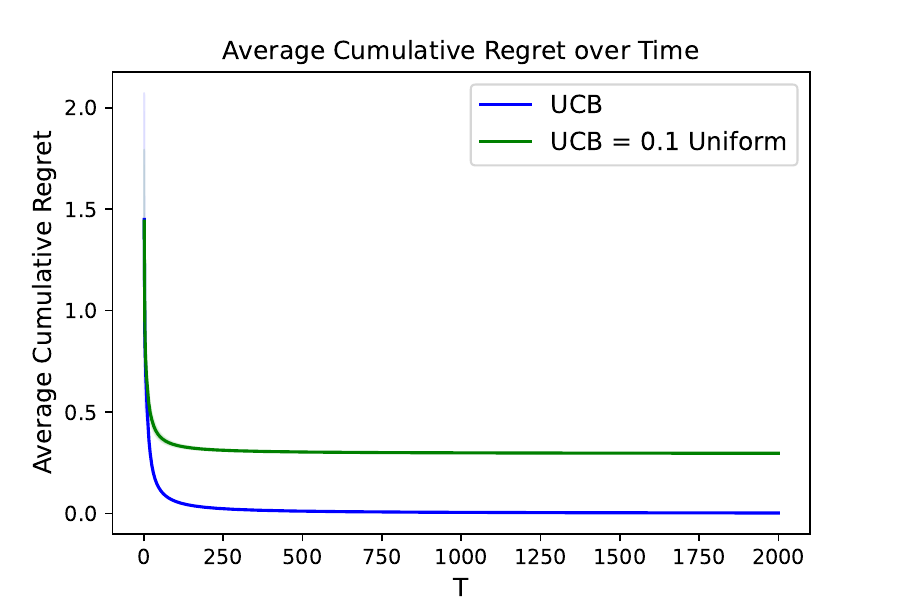}
    \includegraphics[width = 0.35\textwidth]{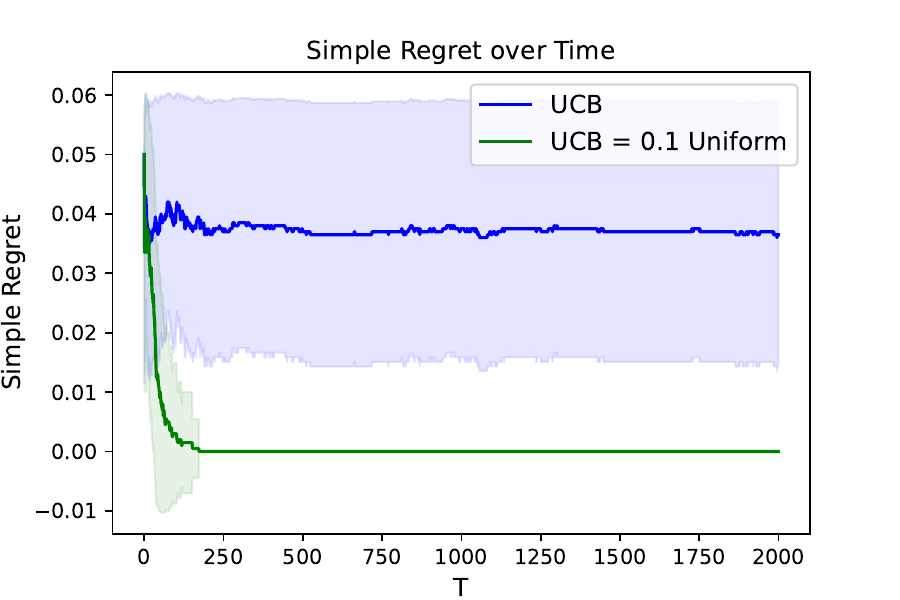}
    \caption{Experiment with changes in policy spaces}
\end{minipage}
\begin{minipage}[b]{1\textwidth}
\centering
    \includegraphics[width = 0.35\textwidth]{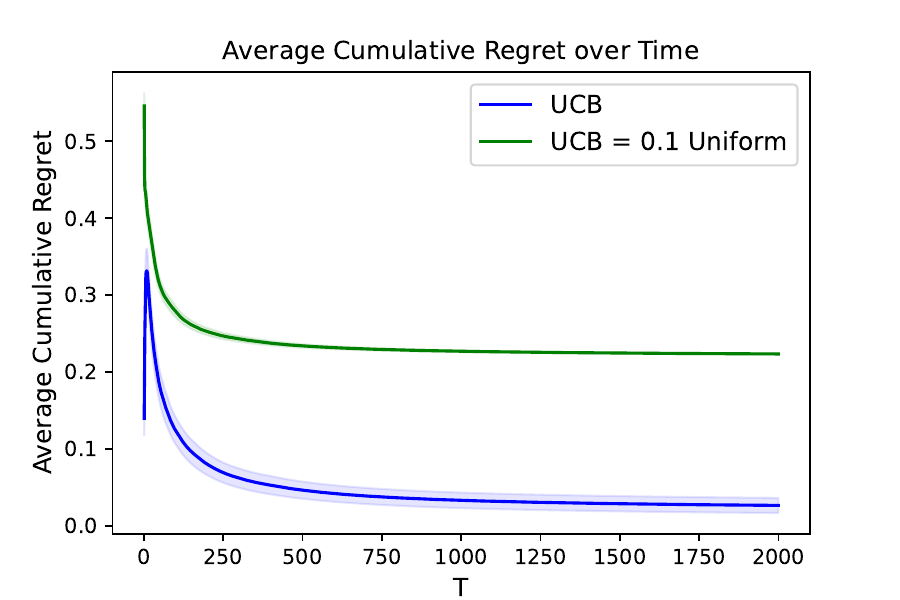}
    \includegraphics[width = 0.35\textwidth]{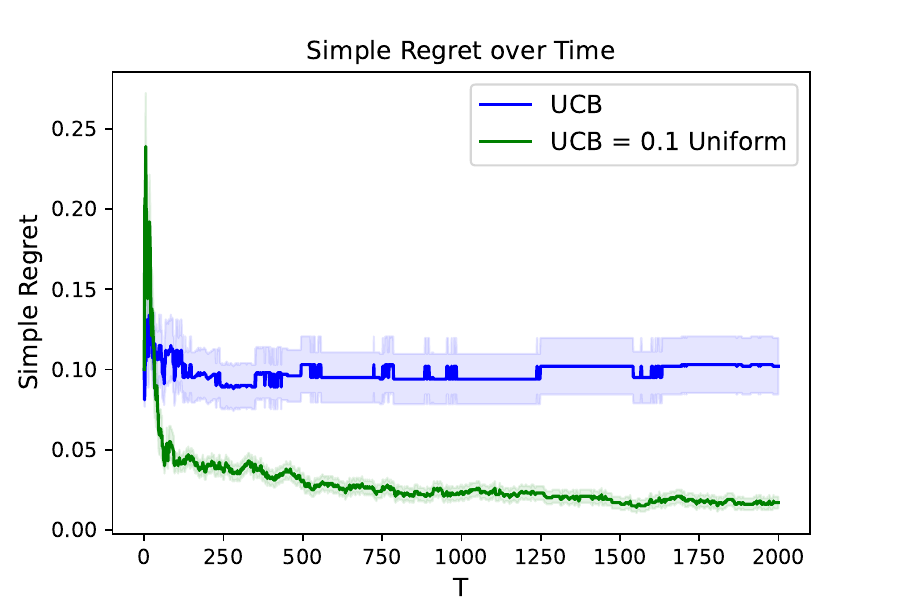}
    \caption{Experiment with changes in reward mappings}
\end{minipage}
  \label{fig:simulation}
\end{figure}
 

In both experiments, UCB shows a diminishing average cumulative regret with a constant simple regret for the second task. Mixed with 0.1 random exploration, UCB receives a constant average regret, while the simple regret goes to zero, revealing a strong trade-off between cumulative regrets and simple regret.

%% file: sections/appendix.tex
\section{Proof Theoerm \ref{thm:1}}

\subsection{Lower Bound When $\Pi^{(1)} \neq \Pi^{(2)}, f^{(1)} = f^{(2)}$}

Recall that our lower bound construction is a set of two-armed contextual bandits with context space $\gX = \{x_1, x_2\}$ and outcome space $\gY = \sR$. We let $\Pi^{(1)} = \{\pi: \pi(\cdot \mid x_1) = \pi(\cdot \mid x_2), \text{for all } x_1, x_2 \in \gX\}$, and $\Pi^{(2)}$ be the set of all policies. We consider $f^{(1)}(y) = f^{(2)}(y) = y$. That is, the reward mappings are identical mappings. Additionally, $P_X$ is uniform distribution over $\gX$.

We consider the instance set (outcome distribution set) $\gP$ such that each $P(\cdot \mid x, a)$ is a Gaussian distribution with mean $p_{x, a}$ and variance $\sigma^2$. Specifically, we consider $P$ such that the mean reward for each context-arm pair is given by Table \ref{tab:case1} for any $\epsilon \in [0, 0.5]$ and $\xi \in [0, 0.25]$. Note that each $\epsilon, \xi$ realizes two instances, denoted by $P_{\epsilon, \xi}$, $\bar P_{\epsilon, \xi}$.

\begin{table}[hpt]
    \centering
    \caption{Case of $\Pi^{(1)} \neq \Pi^{(2)}$}
    \label{tab:case1}
    \begin{tabular}[t]{c|c|c|c || c|c|c|c}
        $P_{\epsilon}$ & $x_1$ & $x_2$ & marginal & $\bar{P}_{\epsilon}$ & $x_1$ & $x_2$ & marginal \\
        \hline
        $a_1$  & $0.5+\epsilon$ & $0.5-\epsilon$ & 0.5 & $a_1$ & $0.5+\epsilon$ & $0.5-\epsilon$ & 0.5  \\
        $a_2$ & $0.5-2\xi$ & 0.5 & $0.5-\xi$ & $a_2$ & $0.5-2\xi$ & $0.5-2\epsilon$ & $0.5-\xi-\epsilon$
\end{tabular}
\end{table}



Now we are ready to prove the theorem. Throughout the proof, we use $\E_P^L[]$ for the expectation of random variables of interest given the underlying instance $P$. Let $T(x, a) = \sum_{t=1}^T \mathbbm{1}\{(X_t, A_t) = (x, a)\}$ the random number of visit in the context-arm pair $(x, a)$.

We first note that 
\begin{align}
    \E^{L^{(1)}}_{P_{\epsilon, \xi}}[\CR] 
    =& \xi \sum_{t= 1}^{T} \E^{L^{(1)}}_{P_{\epsilon, \xi}}[\mathbbm{1}\{A_t = a_2\}]\\
    =& 2\xi\sum_{t= 1}^{T} \E^{L^{(1)}}_{P_{\epsilon, \xi}}[\mathbbm{1}\{A_t = a_2\}] \E^{L^{(1)}}_{P_{\epsilon, \xi}}[\mathbbm{1}\{X_t = x_2\}]\\
    =& 2\xi\sum_{t= 1}^{T} \E^{L^{(1)}}_{P_{\epsilon, \xi}}[\mathbbm{1}\{A_t = a_2\} \mathbbm{1}\{X_t = x_2\}]\\
    =& 2\xi\E^{L^{(1)}}_{P_{\epsilon, \xi}}[T(x_2, a_2)], \label{equ:tmp2}
\end{align}
where the second equality is based on the fact that $X_t \sim \operatorname{Unif}(\gX)$, and the third equality is due to that fact that $A_t \perp X_t$ because $\pi_t \in \Pi^{(1)}$ is context-independent.

For any $\epsilon$, we further provide a lower bound on the sum of squared simple regret of $\hat \pi$. Consider a fixed online learning algorithm $L^{(1)}$.
\begin{align}
    \inf_{L^{(2)}} \sup_{P \in \gP} \E_P^{L}[\SR(\hat \pi)] 
    &\geq \frac{1}{2} \inf_{L^{(2)}} \left(\E^{L}_{P_{\epsilon, \xi}}[\SR(\hat \pi)] + \E^{L}_{\bar P_{\epsilon, \xi}}[\SR(\hat \pi)] \right) \\
    &\geq \frac{\epsilon}{2} \inf_{L^{(2)}} \left( \E^{L}_{P_{\epsilon, \xi}}[1-\mu_{\hat \pi}(x_{2}, a_2)] +  \E^{L}_{P_{\epsilon, \xi}}[\mu_{\hat \pi}(x_{2}, a_2)]\right) \\
    &\geq  \frac{\epsilon}{4} \inf_{L^{(2)}} \left( \sP^{L}_{P_{\epsilon, \xi}} \left( \mu_{\hat \pi}(x_2, a_2) \leq \frac{1}{2} \right) + \sP^{L}_{\bar P_{\epsilon, \xi}} \left( \mu_{\hat \pi}(x_2, a_2) > \frac{1}{2} \right) \right), \label{equ:tmp1}
\end{align}
where the second inequality is based on the fact that $a_1$ is an $\epsilon$-suboptimal arm for $P_{\epsilon, \xi}$, while $a_2$ is an $\epsilon$-suboptimal arm for $\bar{P}_{\epsilon, \xi}$.

\begin{lem}[Bretagnolle–Huber inequality]
\label{lem:BH_inequ}
    For any two probability distributions $P, Q$ on the same measurable space $(\gX, \gF)$, and any event $A \in \gF$, we have
    $$
        P(A) + Q(\bar{A}) \geq \frac{1}{2} \exp(-D_{KL}(P \| Q)).
    $$
\end{lem}

It follows from Lemma \ref{lem:BH_inequ} and the fact that $P_{\epsilon, \xi}$, $\bar{P}_{\epsilon, \xi}$ only differs in $(x_2, a_2)$ that 
\begin{align}
    &\ \inf_{L^{(2)}} \left( \sP^{L}_{P_{\epsilon, \xi}} \left( \mu_{\hat \pi}(x_2, a_2) \leq \frac{1}{2} \right) + \sP^{L}_{\bar P_{\epsilon, \xi}} \left( \mu_{\hat \pi}(x_2, a_2) > \frac{1}{2} \right) \right) \\
    \geq& \frac{1}{2} \exp(- \E^{L^{(1)}}_{P_{\epsilon, \xi}}[T(x_2, a_2)] D_{KL}(P_{\epsilon, \xi} \| \bar{P}_{\epsilon, \xi})) \\
    =& \frac{1}{2} \exp\left(- \frac{1}{2}\E^{L^{(1)}}_{P_{\epsilon, \xi}}[T(x_2, a_2)]  \log\left(\frac{1}{1-4\epsilon^2} \right)\right) \\
    \geq& \frac{1}{2} \exp\left(- 2\E^{L^{(1)}}_{P_{\epsilon, \xi}}[T(x_2, a_2)] \epsilon^2\right)
\end{align}
where the second inequality holds because $\log(1/(1-x)) \geq x$ for any $x \in [0, 1/2]$.

Combined with (\ref{equ:tmp1}), we have for any given online learning algorithm $L^{(1)}$, and any $\xi \in [0, 0.25]$,
\begin{align}
    \inf_{L^{(2)}} \sup_{P \in \gP} \E_P^{L}[\SR(\hat \pi)] \geq \frac{\epsilon}{8} \exp(-2\epsilon^2 \E^{L^{(1)}}_{P_{\epsilon, \xi}}[T(x_2, a_2)]) = \frac{\epsilon}{8} \exp(-2\epsilon^2 \E^{L^{(1)}}_{P_{0, \xi}}[T(x_2, a_2)]). \label{equ:tmp2}
\end{align}

The second equality holds because $\E_{P_{\epsilon, \xi}}^{L^{(1)}}[\cdot] = \E_{P_{0, \xi}}^{L^{(1)}}[\cdot]$ for any $\epsilon$ due to the fact that the online learning algorithm in task 1 learns a multi-armed bandit (MAB), and we have the same (MAB) for all $\epsilon$ because the marginal distribution $(P(\cdot \mid x_1, a) + P(\cdot \mid x_2, a))/2$ is independent of $\epsilon$.

Combined with (\ref{equ:tmp2}), we have 
\begin{equation}
    \inf_{L^{(2)}} \sup_{P \in \gP} \E_P^{L}[\SR(\hat \pi)] \geq \frac{\epsilon}{8} \exp(-\epsilon^2 \E^{L^{(1)}}_{P_{0, \xi}}[\CR] / \xi). \label{equ:tmp3}
\end{equation}

To finish the proof, we follow a similar argument in \cite{simchi2023multi}. We let $\hat P_L = \argmax_{P \in \gP} \E_P^L[\CR]$, and $\epsilon = \sqrt{\xi / \E_{\hat P_L}^{L}[\CR]}$. By symmetry $P_{0, \xi}$ and $\bar{P}_{0, \xi}$ represents the same space, we know that $\E_{\hat P_L}^{L}[\CR] = \E_{P_{0, \xi}}^{L}[\CR]$ for some $\xi$. Then for any $L$, we apply (\ref{equ:tmp3})
\begin{equation}
    \sup_{P \in \gP} \E_{P}^{L}[\SR] \geq \frac{\epsilon}{8} \exp(-\epsilon^2 \E^{L}_{P_{0, \xi}}[\CR] / \xi) \geq \epsilon / (8e).
\end{equation}

Plugging in $\epsilon$, we have 
\begin{align}
    \sup_{P \in \gP} \left[\E_{P}^L [\SR] \sqrt{\E_P^L[\CR]} \right] 
    &\geq \left( \E_{\hat P_L}^L[\SR] \sqrt{\E_{\hat P_L}^L[\CR]} \right) \\
    &= \left(\frac{\epsilon}{8e} \sqrt{\E_{\hat P_L}^L[\CR]} \right)\\
    &= \Theta(1).
\end{align}

This argument holds for any choice of $L$, which completes the proof.

\subsection{Lower Bound When $f^{(1)} = f^{(2)}, \Pi^{(1)} = \Pi^{(2)}$}

The proof for the second statement in Theorem \ref{thm:1} relies on the lower bound construction where the policy space $\Pi$ remains the same. Recall that we construct a multi-armed bandit environment without context. The outcome vector is a two-dimensional vector $Y_{t} = (Y_{t, 1}, Y_{t, 2})$. Table \ref{tab:case2_main_full} specifies a set of outcome distributions parametrized by $\epsilon, \xi$. Each pair of $\epsilon, \xi$ determines the mean rewards for $\E[Y_{t, i} \mid A_t = a_j]$ for all $i, j \in \{1, 2\}$.
\begin{table}[H]
    \caption{Mean reward for $f^{(1)} \neq f^{(2)}$ case}
    \label{tab:case2_main_full}
    \centering
    \begin{tabular}[t]{c|c|c || c | c|c}
        $P_{\epsilon, \xi}$ & $Y_{t, 1}$ & $Y_{t, 2}$ & $\bar P_{\epsilon, \xi}$ & $Y_{t, 1}$ & $Y_{t, 2}$ \\
        \hline
        $a_1$  & 0.5 & 0.5 & $a_1$  & 0.5 & 0.5\\
        $a_2$  & 0.5-$\xi$ & 0.5-$\epsilon$  & $a_2$ & 0.5-$\xi$ & 0.5+$\epsilon$
    \end{tabular}
\end{table}

Following a similar proof in the previous section, we let $T(a) = \sum_{t=1}^T \mathbbm{1}\{A_t = a\}$ be the total number of pulls of $a$ in task 1.

We first note that 
\begin{align}
    \E^{L^{(1)}}_{P_{\epsilon, \xi}}[\CR] 
    = \xi \sum_{t= 1}^{T} \E^{L^{(1)}}_{P_{\epsilon, \xi}}[\mathbbm{1}\{A_t = a_2\}] = \xi \E^{L^{(1)}}_{P_{\epsilon, \xi}}[T(a_2)]. \label{equ:tmp2}
\end{align}

For any $\epsilon$, we further provide a lower bound on the sum of squared simple regret of $\hat \pi$. Consider a fixed online learning algorithm $L^{(1)}$.
\begin{align}
    \inf_{L^{(2)}} \sup_{P \in \gP} \E_P^{L}[\SR(\hat \pi)] 
    &\geq \frac{1}{2} \inf_{L^{(2)}} \left(\E^{L}_{P_{\epsilon, \xi}}[\SR(\hat \pi)] + \E^{L}_{\bar P_{\epsilon, \xi}}[\SR(\hat \pi)] \right) \\
    &\geq \frac{\epsilon}{2} \inf_{L^{(2)}} \left( \E^{L}_{P_{\epsilon, \xi}}[1-\mu_{\hat \pi}(a_2)] +  \E^{L}_{P_{\epsilon, \xi}}[\mu_{\hat \pi}(a_2)]\right) \\
    &\geq  \frac{\epsilon}{4} \inf_{L^{(2)}} \left( \sP^{L}_{P_{\epsilon, \xi}} \left( \mu_{\hat \pi}(a_2) \leq \frac{1}{2} \right) + \sP^{L}_{\bar P_{\epsilon, \xi}} \left( \mu_{\hat \pi}(a_2) > \frac{1}{2} \right) \right), \label{equ:tmp1_2}
\end{align}
where the second inequality is based on the fact that $a_1$ is an $\epsilon$-suboptimal arm for $P_{\epsilon, \xi}$, while $a_2$ is an $\epsilon$-suboptimal arm for $\bar{P}_{\epsilon, \xi}$.

It follows from Lemma \ref{lem:BH_inequ} and the fact that $P_{\epsilon, \xi}$, $\bar{P}_{\epsilon, \xi}$ only differs in $a_2$ that 
\begin{align}
    &\ \inf_{L^{(2)}} \left( \sP^{L}_{P_{\epsilon, \xi}} \left( \mu_{\hat \pi}(a_2) \leq \frac{1}{2} \right) + \sP^{L}_{\bar P_{\epsilon, \xi}} \left( \mu_{\hat \pi}(a_2) > \frac{1}{2} \right) \right) \\
    \geq& \frac{1}{2} \exp(- \E^{L^{(1)}}_{P_{\epsilon, \xi}}[T(a_2)] D_{KL}(P_{\epsilon, \xi} \| \bar{P}_{\epsilon, \xi})) \\
    =& \frac{1}{2} \exp\left(- \frac{1}{2}\E^{L^{(1)}}_{P_{\epsilon, \xi}}[T(a_2)]  \log\left(\frac{1}{1-4\epsilon^2} \right)\right) \\
    \geq& \frac{1}{2} \exp\left(- 2\E^{L^{(1)}}_{P_{\epsilon, \xi}}[T(a_2)] \epsilon^2\right)
\end{align}
where the second inequality holds because $\log(1/(1-x)) \geq x$ for any $x \in [0, 1/2]$.

Combined with (\ref{equ:tmp1_2}), we have for any given online learning algorithm $L^{(1)}$, and any $\epsilon, \xi \in [0, 0.25]$,
\begin{align}
    \inf_{L^{(2)}} \sup_{P \in \gP} \E_P^{L}[\SR(\hat \pi)] \geq \frac{\epsilon}{8} \exp(-2\epsilon^2 \E^{L^{(1)}}_{P_{\epsilon, \xi}}[T(a_2)]). \label{equ:tmp2_2}
\end{align}

By symmetry, there exists some $\bar{P}_{0, \xi}$ such that $\E^{L^{(1)}}_{P_{\epsilon, \xi}}[T(a_2)] = \E^{L^{(1)}}_{\bar P_{0, \xi}}[T(a_2)]$. To finish the proof, we follow a similar argument in \cite{simchi2023multi}. We let $\hat P_L = \argmax_{P \in \gP} \E_P^L[\CR]$, and $\epsilon = \sqrt{\xi / \E_{\hat P_L}^{L}[\CR]}$. By symmetry $\bar{P}_{0, \xi}$ represents the whole set of $\gP$, we know that $\E_{\hat P_L}^{L}[\CR] = \E_{P_{0, \xi}}^{L}[\CR]$ for some $\xi$. Then for any $L$, we apply (\ref{equ:tmp3})
\begin{equation}
    \sup_{P \in \gP} \E_{P}^{L}[\SR] \geq \frac{\epsilon}{8} \exp(-\epsilon^2 \E^{L}_{\bar{P}_{0, \xi}}[\CR] / \xi) \geq \epsilon / (8e).
\end{equation}

Plugging in $\epsilon$, we have 
\begin{align}
    \sup_{P \in \gP} \left[\E_{P}^L [\SR] \sqrt{\E_P^L[\CR]} \right] 
    &\geq \left( \E_{\hat P_L}^L[\SR] \sqrt{\E_{\hat P_L}^L[\CR]} \right) \\
    &= \left(\frac{\epsilon}{8e} \sqrt{\E_{\hat P_L}^L[\CR]} \right)\\
    &= \Theta(1).
\end{align}

This argument holds for any choice of $L$, which completes the proof.

\section{Proof of Proposition \ref{prop:lower_bound_sum}.}

\textbf{Proposition \ref{prop:lower_bound_sum}.} 
\textit{
Following the same choice of the instance set $\gP$ and $\Pi^{(1)}, \Pi^{(2)}, f^{(1)}, f^{(2)}$ in Theorem \ref{thm:1}, the following minimax lower bound holds
\begin{equation}
    \inf_{L \in \gL} \sup_{P \in \gP} \E[\CR + T'\SR] = \Omega\left( \max\left\{\frac{T'}{\sqrt{T}}, T'^{2/3}, \sqrt{T}\right\}\right).\label{equ:three_lower}
\end{equation}}

\begin{proof}
    Since $\inf_{L \in \gL} \sup_{P \in \gP} \sqrt{\E[\CR]} \times \E[\SR] = \Theta(1)$,  we have
    \begin{align}
        \inf_{L \in \gL} \sup_{P \in \gP} \E[\CR + T'\SR] 
        = \Omega\left(\inf_{L \in \gL} \sup_{P \in \gP} \frac{1}{(\E[\SR])^2}+\E[T'\SR] \right)
        = \Omega\left( T'^{2/3} \right).
    \end{align}
    The well-established cumulative regret lower bound \citep{lattimore2020bandit} gives us
    \begin{equation}
        \inf_{L \in \gL} \sup_{P \in \gP} \E[\CR] =  \Omega(\sqrt{T}).
    \end{equation}
    An information-theoretic lower bound on the estimation error gives us 
    \begin{equation}
        \inf_{L \in \gL} \sup_{P \in \gP} T'\E[\SR] = \Omega(T' / \sqrt{T}).
    \end{equation}
    The proof is finished by combining the above three lower bounds.
\end{proof}

\section{Proof of Theorem \ref{thm:upper_bound}}
\textbf{Theorem \ref{thm:upper_bound}.}
\textit{
    Let $L_0$ be an online learning algorithm with a regret bound of $\tilde{\gO}(\sqrt{|\gX||\gA|T})$. Let the algorithm for the first task be $L_{\alpha}(\tau) = (1-\alpha) L_0(\tau) + \alpha \pi_0$, where $\tau$ is any past observations and $\pi_0$ is the uniform random policy. For any choice of $\alpha \in [{|\gX| |\gA|}/{\sqrt{T}}, 1]$, there exist offline-learning algorithm for the second task such that
    \begin{equation}
        \E[\CR] = \tilde{\gO}\left(\alpha T \right), \E[\SR] = \tilde{\gO}\left(\sqrt{\frac{(|\gX| |\gA|)^2}{\alpha T}} \right). \label{equ:upper_bound}
    \end{equation}
}

\begin{proof}
    The upper bound is straightforward from
    $$
        \E[\CR] = \tilde{O}(\sqrt{|\gX|  |\gA| T} + \alpha T) = \tilde{O}(\alpha T),
    $$
    where the second inequality is due to the fact that $\alpha T \geq {|\gX| |\gA| \sqrt{T}}$. To establish a valid simple regret upper bound, we employ an importance weight estimator:
    \begin{equation}
        \hat R(x, a) \coloneqq \frac{\sum_{t = 1}^T \left[ \mathbbm{1}\{(X_{t}, A_t) = (x, a)\} \frac{\pi_0(A_t \mid X_t)}{L_{\alpha}(\tau_t)(A_t \mid X_t)} f^{(2)}(Y_t)  \right]}{\sum_{t = 1}^T \left[ \mathbbm{1}\{(X_{t}, A_t) = (x, a)\} \frac{\pi_0(A_t \mid X_t)}{L_{\alpha}(\tau_t)(A_t \mid X_t)}  \right]}.
    \end{equation}
    Theorem 2.1. \citep{guo2024online}  states that it holds for all $x, a$ that
    \begin{equation}
        |\hat R(x, a) - R(x, a)| = \tilde{O}\left(\sqrt{\frac{|\gX||\gA|^2}{\alpha T}} \right).
    \end{equation}
    Then the policy $\hat \pi$ defined by
    \begin{equation}
        \hat \pi(a \mid x) = \operatorname{Unif}(\{\argmax_{a' \in \gA} \hat R(x, a')\})
    \end{equation}
    achieves the simple regret bound of $\tilde{O}(|\gX| |\gA| / \sqrt{\alpha T})$.
\end{proof}

\section{Proof of Theorem \ref{thm:lower_bound_p}}

{\bf Theorem \ref{thm:lower_bound_p}.}  {\it Assume $\gS$ is such that $\Pi^{(1)} = \Pi^{(2)} = \Pi$, the set of all policies, and $\Pi^{(1)} = \Pi^{(2)} = f$, the identical mapping in $\sR$. Assume each $P^{(i)}(\cdot \mid x, a)$ is from a binomial distribution with mean $P^{(i)}(x, a)$ for all $i = 1, 2$ and $x, a \in \gX \times \gA$, and $P^{(2)} \in \gP(P^{(1)}, B)$. Then there exists some $B$ such that 
\begin{equation}
    \inf_{L \in \gL_2} \sup_{P^{(1)}} \sqrt{\E[\Reg_1]} \E[\widetilde{\SR}(\pi_2 \mid P^{(1)}, B)] = \Omega(1),
\end{equation}
where $\pi_2$ is the random policy chosen by learning algorithm $L$.
}

\begin{proof}
    We construct two hard instances inspired by Proposition \ref{prop:simple-regret}. Recall that Proposition \ref{prop:simple-regret} states 
    that any two-armed, context-free bandit, with $\Gap \coloneqq P^{(1)}(a_1) - P^{(1)}(a_2) > 0$ has the following explicit form of optimal robust policy:
    \begin{equation}
        \tilde{\pi}(a_1) = \frac{B + \Gap}{2B}, \text{ and } \tilde{\pi}(a_2) = \frac{B - \Gap}{2B}.
    \end{equation}

Let the arm space be $\gA = \{a_1, a_2\}$. We construct two instances $S$ and $\bar{S}$ with $P^{(1)}$ and $\bar{P}^{(1)}$, such that $R^{(1)}(a_1) = \bar{R}^{(1)}(a_1) = 1$ and $R^{(1)}(a_2) = 1/2$, while $\bar{R}^{(1)}(a_2) = 1/2 + \epsilon$ for $\epsilon < 1/4$. Let $P^{(i)}(\cdot \mid a)$ and $\bar{P}^{(i)}(\cdot \mid a)$ be Bernoulli distributions of parameter $R^{(i)}(a)$ and $\bar{R}^{(i)}(a)$ for each $i \in \{1, 2\}, a \in \gA$. Let $\Gap = R^{(1)}(a_1) - R^{(1)}(a_2) = 1/2$ and $\overline{\Gap} = \bar{R}^{(1)}(a_1) - \bar{R}^{(1)}(a_2) = 1/2 - \epsilon$.

We first connect the cumulative regret in the first task $\Reg_1$ with the number of visits in the suboptimal arm $a_2$ in the first task. It can be shown that $\E_{\mS'}[\Reg_1] \geq 1/4 \E_{\mS'}[T(a_2)]$, where $T(a_2) \coloneqq \sum_{t = 1}^{T_1} \mathbbm{1}_{A_{1, t} = a_2}$ for each $S' \in \{S, \bar{S}\}$.

We consider $B = 3/4 > \max\{\Gap, \overline{\Gap}\}$. By Proposition \ref{prop:simple-regret}, we first lower bound the robust simple regret by
\begin{align}
        & \widetilde{\SR}(\pi \mid P^{(1)}, B) \\
    =   & \SR(\pi \mid P^{(1)}, B) - \inf_{\pi'}\SR(\pi' \mid P^{(1)}, B) \\
    =   & \max\{(B - \text{Gap})\pi(a_1), (B - \text{Gap})\pi(a_2)\} - \frac{B^2 - \Gap^2}{2B}\\
    =   & \left(\pi(a_1) - \frac{B + \Gap}{2B} \right)^{+}(B - \Gap) + \left(\pi(a_2) - \frac{B - \Gap}{2B} \right)^{+}(B + \Gap)\\
    \geq& (B - \Gap)|\pi - \pi^*|/2 \geq |\pi - \pi^*| / 8.
\end{align}
Similarly, we also have $\widetilde{\SR}(\pi \mid \bar P^{(1)}, B) \geq |\pi - \bar{\pi}^*|/8$. Here we let $\pi^*$, $\bar{\pi}^*$ be the optimal robust policy for $P^{(2)} \in \gP(P^{(1)} \mid B)$ and $P^{(2)} \in \gP(\bar{P}^{(1)} \mid B)$, respectively.

Let $\pi_2$ be the random policy proposed by the learning algorithm for the second task. We convert the robust learning problem to a testing problem of two instances. Note that $\pi^*(a_1) = 5/6$ and $\bar{\pi}^*(a_1) = 5/6 - 2\epsilon/3$.

The sum of robust simple regrets for two instances can be lower bounded by
\begin{align}
        & \E_{P^{(1)}}[\widetilde{\SR}_{2}] + \E_{\bar{P}^{(1)}}[\widetilde{\SR}_{2}] \\
    \geq& \frac{\epsilon}{24} \sP_{P^{(1)}}\left(\pi_2(a_1) \leq \frac{5}{6} - \epsilon/3 \right) + \frac{\epsilon}{24} \sP_{\bar{P}^{(1)}}\left(\pi_2(a_1) > \frac{5}{6} - \epsilon/3 \right) \\
    \geq& \frac{\epsilon}{24} \exp(-\epsilon^2 \E_{P^{(1)}}[T(a_2)])
\end{align}

Choosing $\epsilon = \sqrt{1/\E_{P^{(1)}}[T(a_2)]}$ and applying Lemma \ref{lem:BH_inequ} again, we have 
$$
    \inf_{L \in \gL} \sup_{P^{(1)}} \sqrt{\E[\Reg_1]} \E[\widetilde{\SR}(\pi_2 \mid P^{(1)}, B)] = \Omega(1).
$$
\end{proof}






\section{Proof of Theorem \ref{thm:UCB}}

\textbf{Theorem \ref{thm:UCB} }{\it
    Consider the hypothesis set $\gF$ to be one-hidden layer neural networks with width $d$. There exists ground-truth reward function and a sequence of tasks of length $\Omega(\exp(d))$ with different $\Pi^{(i)}$, such that the cumulative regret for each task is lower bounded by a constant, even if each $T_i \rightarrow \infty$.
}

\begin{proof}
    The construction of the hard instance can be described below. Consider a nonlinear bandit problem with $\gA = S^{d-1}$, the $d$-dimensional sphere. We first define the reward function as 
    $$
        R(\theta_2) = \alpha_1 \langle \theta_1, a \rangle  + \alpha_2\left( \langle \theta_2, a \rangle - \epsilon \right)^+,
    $$
    where $\theta_1 \in S^{d-1}, \alpha_1 > 0$ and $\alpha_2 > 0$ are known parameters and $\theta_2 \in S^{d-1}$ is unknown. Assume that the true parameter $\theta_2^*$ satisfies $\langle \theta_1, \theta_2^* \rangle < 0$, i.e., $\theta_1$ and $\theta_2^*$ are on different sphere. Furthermore, let $\alpha_2 = 2\alpha_2 / (1-\epsilon)$.

    Let $\{\gA_1, \dots, \gA_N\}$ be an $\epsilon$-pack of the subset $\{a \in \gA: \langle \theta_1, a \rangle < 0\}$. Let the allowed policy space for task $i$ be $\Pi^{(i)} = \{\pi: \pi \text{ is supported on } \gA_i \cup \{\theta_1\}\}$. Specifically, order $\{\gA_1, \dots, \gA_N\}$ such that $\theta_2^* \in \gA_{N}$.

    To verify, $R(\theta_2)$ is in the family of one-layer neural network with ReLU activation function.

    We first observe that the optimal policy for task $i$ is $\pi^*_i = \delta(\theta_1)$ for all $i = 1, \dots N-1$. Note that by running UCB, the algorithm will optimistically choose $a \in \gA_i$ as they do not know that whether $\theta_2^* \in \gA_i$ for all $i = 1, \dots, N-1$, thus leading to a constant regret for all tasks $i < N$.
\end{proof}